\documentclass[sigconf,nonacm]{acmart}
\usepackage[utf8]{inputenc}
\usepackage[ruled, vlined, linesnumbered]{algorithm2e}
\usepackage[scaled=.78]{beramono}
\usepackage{bm}
\usepackage[labelfont={small,bf}, textfont={small}]{caption}
\usepackage{enumitem}
\usepackage{hhline}
\usepackage{multirow}
\usepackage{makecell}
\usepackage{rotating}
\usepackage{subfigure}
\usepackage{tikz}
\def\checkmark{\tikz\fill[scale=0.3](0,.35) -- (.25,0) -- (1,.7) -- (.25,.15) -- cycle;}
\usepackage{xspace}

\newcommand\blfootnote[1]{ 
  \begingroup
  \renewcommand\thefootnote{}\footnote{#1} 
  \addtocounter{footnote}{-1} 
  \endgroup
}

\makeatletter
\g@addto@macro\bfseries{\boldmath}
\makeatother

\newcommand{\overbar}[1] {\mkern 1.5mu\overline{\mkern-3.5mu#1\mkern-1.0mu}\mkern 1.5mu}

\newcommand{\agreement}{\mathcal{A}}
\newcommand{\performance}{\mathcal{M}} 

\newcommand{\Sec}[1]		{Sec.\,\ref{#1}}
\newcommand{\Fig}[1]		{Fig.\,\ref{#1}}

\newcommand{\Eq}[1]			{Eq.\,\ref{#1}}
\newcommand{\Tab}[1]		{Tab.\,\ref{#1}}
\newcommand{\Alg}[1]		{Alg.\,\ref{#1}}

\newcommand{\ie}   			{i.e.\@\xspace}
\newcommand{\eg}   			{e.g.\@\xspace}

\newcommand{\mydots} 	{...}
\newcommand{\dset}[1]  {\texttt{#1}}

\newcommand{\inlinetitle}[2]  {\vspace{4pt}\noindent\textbf{\emph{#1}{#2}}}

\DeclareMathOperator*{\argmax}{\arg\!\max}

\begin{document}

\title[Model family selection for classification using Neural Decision Trees]{Model family selection for classification \\using Neural Decision Trees}

\author{Anthea Mérida Montes de Oca \qquad Argyris Kalogeratos$^{*}$ \qquad Mathilde Mougeot}

\authornote{Corresponding author}

\affiliation{
\vspace{2mm}
\institution{Centre Borelli, ENS Paris-Saclay, Universite Paris-Saclay, FR-94230 Cachan, France}
\vspace{4mm}
}

\renewcommand{\shortauthors}{Mérida, et al.}

\begin{abstract}
Model selection consists in comparing several candidate models according to a metric to be optimized. The process often involves a grid search, or such, and cross-validation, which can be time consuming, as well as not providing much information about the dataset itself. In this paper we propose a method to reduce the scope of exploration needed for the task. The idea is to quantify how much it would be necessary to depart from \emph{trained instances of a given family}, reference models (RMs) carrying `rigid' decision boundaries (\eg decision trees), so as to obtain an equivalent or better model. In our approach, this is realized by progressively relaxing the decision boundaries of the initial decision trees (the RMs) as long as this is beneficial in terms of performance measured on an analyzed dataset. More specifically, this relaxation is performed by making use of a neural decision tree, which is a neural network built from DTs. The final model produced by our method carries non-linear decision boundaries. Measuring the performance of the final model, and its agreement to its seeding RM can help the user to figure out on which family of models he should focus on. 

\end{abstract}

\keywords{Model selection, decision trees, neural decision trees, neural networks,
model family selection,
interpretability}

\maketitle

\section{Introduction}\label{sec:intro}
\blfootnote{
Authors' emails:
{\tt\footnotesize  \{merida,\,kalogeratos,\,mougeot\}@cmla.ens-cachan.fr}.\\
This work was funded by the IdAML Chair hosted at ENS Paris-Saclay.
}
For both cultures in modeling
inference and prediction \cite{Breiman2001}, model selection and model procedure selection are essential steps to obtain a  satisfactory solution to a problem at hand, and hence a wide variety of methods and metrics \cite{Ding2018} 
do exist to help the user decide which model fits his data and is better to use.  
 
In machine learning, where it is more usual to focus on the prediction ability of models, cross-validation (CV) is a widespread procedure both for model and algorithm selection \cite{Ding2018}.
Indeed, CV has the advantages of being easy to implement and simplifying the comparison of models from different families on the basis of a metric and its variability. It is always necessary, though, to choose which candidate models are to be compared. In this paper, we focus on how to select the model family from which to get candidate models, and specifically to challenge a rigid trained model through relaxation.

\begin{figure}[t]\small
\centering
\includegraphics[width=0.95\columnwidth]{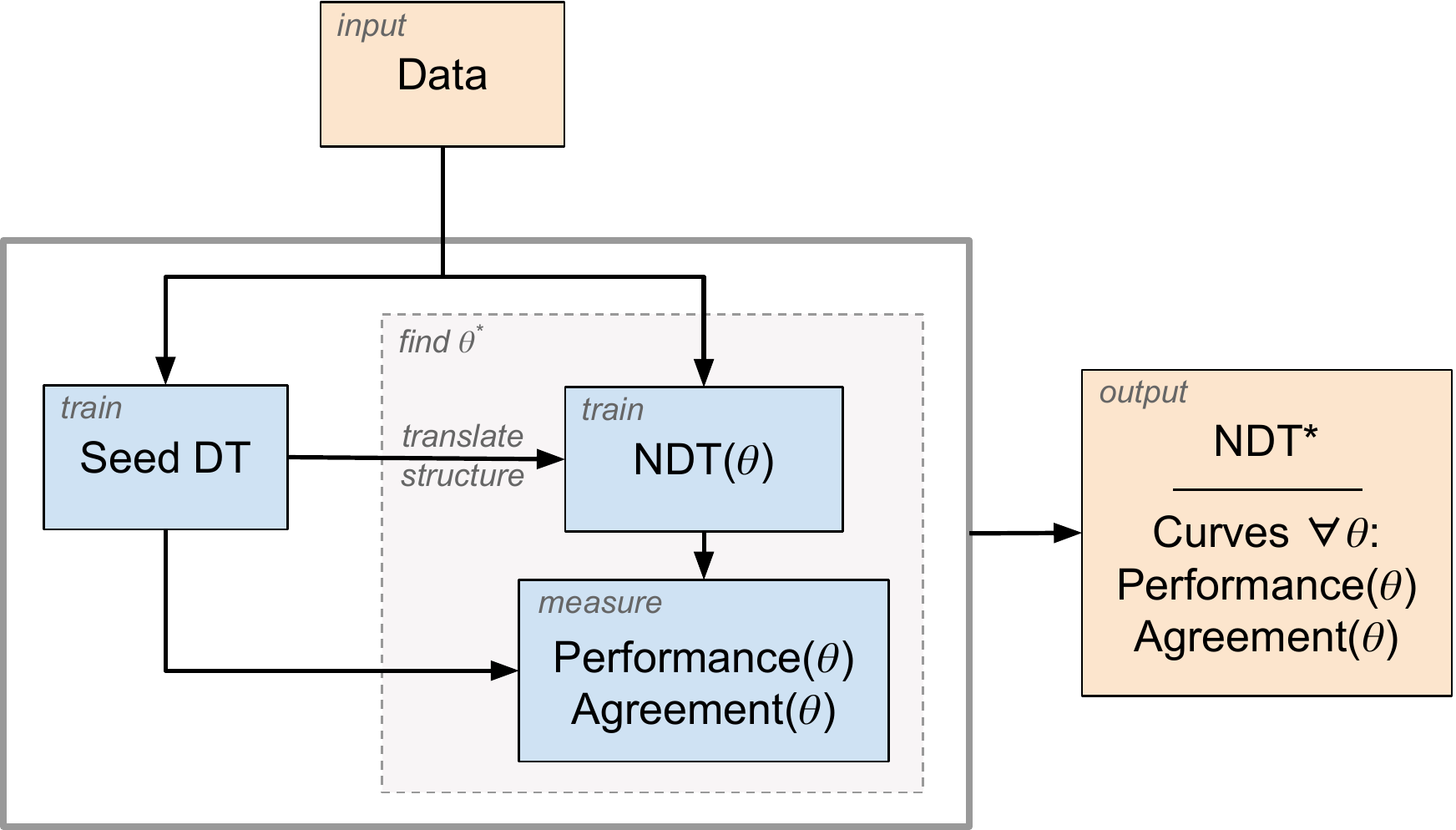}
\vspace{-0.5em}
\caption{Outline of the proposed method.}
\vspace{-1.2em}

\label{fig:outline}
\end{figure}

Currently, there exist packages such as Auto-WEKA \cite{Kotthoff2019} and Auto-sklearn \cite{Feurer2015},
that provide tools to automate the algorithm and model selection process. The methods included in these packages apply Bayesian optimization and meta-learning procedures, while considering the modeling algorithm to be a hyperparameter. In particular, they select the most appropriate algorithm and the values for its parameters according to a metric, and within a budget set by the user. Our approach differs in that its output is not a specific trained model, or even a specific model family
, but a set of indicators that can help the user gain insight regarding the kind of model he needs, as portrayed by \Fig{fig:outline}. The user is thus directly involved in the process of model selection, and this is how our method sets itself apart from existing automated machine learning approaches.

In a similar direction, there are methods (\eg \cite{Ali2006, Smith2002}
that have been developed to obtain insight into what kind of data mining algorithm is suitable to a dataset according to certain 
statistical and information theoretical measures. Both works employ methods that can be interpreted by users: \cite{Ali2006}
generates rules using the C5.0 algorithm and \cite{Smith2002}
clusters datasets according to their characteristics using self-organizing maps and associates each cluster with the algorithms that perform best on average for the datasets lying therein. These methods require a large set of datasets and their evaluation results for different algorithms. The method presented in this paper, however, can be applied directly to a dataset.

Our aim is not to pinpoint to a specific algorithm as the best for a given task, but rather to guide the user when selecting a model family to further explore the most promising one with model selection techniques. More specifically, we propose a methodology that is based on measuring the departure (\ie disagreement at the decisions level) between a rigid reference model and a more flexible, which is initialized by the first one, and whose decision boundaries we relax in a controlled way. 
Here, decision trees (DTs), which are highly interpretable and partition the feature space in hyperrectangles, are confronted to neural networks (NNs) which are capable of approximating non-linear interactions in the feature space. 
The procedure explores the space between DTs and NNs 
by optimizing the parametrization of a neural decision tree (NDT), following the procedure outlined in \Fig{fig:outline}.

\section{Background}\label{sec:background}

\subsection{Neural decision trees (NDTs)} 

NDTs are NNs whose architecture and weights initialization are obtained directly from an input DT, which we call `seed DT'. The variant of NDTs we use here is presented in \cite{Biau2018}. In this case, there is no need to search for the right number of layers or the number of neural units on each layer. 
 
NDTs are always formed by four layers: an input layer, two hidden layers, and an output layer. The connections between the layers encode the information extracted from the seed DT. For a dataset with $d$ features and a seed DT with $K$ leaves, we have the architecture and initialization that we discuss next.

\inlinetitle{First hidden layer}{.} It encodes the information from the inner nodes (or split nodes) of the input DT. It is formed by $K-1$ units, each one representing a split node. The conditional splits in a tree are formed by a feature and a threshold on it. The NDT encodes this information in the weight matrices of the connections between this layer and the input layer: each column in the weight matrix is a one-hot encoding of the feature used in each split and the values of the biases are the opposite values of the thresholds.

\inlinetitle{Second hidden layer}{.} It contains the paths in the DT from the root to the leaves. It consists in $K$ neurons, one for each leaf of the DT. Then, the connections between the units of this layer and those of the previous layer encode the positions of the leaves with respect to each split node. The
elements of the weight matrix take three possible values:

\begin{itemize}[leftmargin=0.6cm]
  \item $-1$ if the leaf is in the path of the inner node and is "on the right side" of the split node;
  \item $1$ if the leaf is in the path of the inner node and is "on the left side" of the split node;
  \item $0$ otherwise.
\end{itemize}
As for the biases, they take the value $-l(k)+\frac{1}{2}$,
where $l(k)$ is the length of the path from the root of the tree to the leaf $k$.

\inlinetitle{Output layer}{.} For a classification task, the NDT outputs the observed probability of an instance falling in the leaf $k$. The values of the weight matrix are $N_k/N$, where $N_k$ is the number of training instances assigned to leaf $k$, and $N$ is the number of training instances. The bias has the value $0$.

\inlinetitle{Activation functions}{.} In a DT, the splits, as well as the leaf memberships, are crisp. In order for an NDT to behave like a DT, its activation functions have to be crisp as well. However, a crisp function is not differentiable so it would not be possible to train the NDT using backpropagation. To mitigate this problem, in \cite{Biau2018} it is proposed to approximate the crisp threshold function of the trees with the function $s$:
\begin{equation}
  \sigma: x \mapsto \tanh(\gamma x).
\end{equation}
The $\gamma$ parameter allows to control the  smoothness of the $\sigma$ function: the higher the value of $\gamma$, the steeper the curve of $\sigma$ gets. Moreover, $\sigma$ is used as an activation function for both of the hidden layers in the NDT, each with a different value of the $\gamma$ parameter, which we denote by $\gamma_1$ for the first hidden layer and by $\gamma_2$ for the second one.

\subsection{Metrics}\label{sec:metrics}

Having obtained a trained NDT that was initialized by a specific input DT, we can compare their performance on the dataset of interest and measure the departure of the second with respect to the first one. For this purpose, we specifically measure the agreement $\agreement$ between the an NDT and its seed DT.
We employ different metrics for classification and regression tasks.

For classification, the agreement metric used is Cohen's $\kappa$ statistic \cite{Cohen1960}, which calculates the extent to which the labels found by two classifiers agree, taking into account the probability for them to agree by chance:
\begin{equation}
  \kappa = \frac{p_o-p_e}{1-p_e},
\end{equation}
where $p_o$ is the probabilities that the two classifiers do agree, and $p_e$ the probability that they agree by chance. When the models agree completely $\kappa = 1$; when they agree by chance $\kappa = 0$; if their agreement is less than what is expected by chance, then $\kappa < 0$.

By abusing the notation, we write as $\agreement(NDT(\gamma))$ the agreement of an NDT with the seed DT that initialized it.

Finally, the metric, denoted by $\performance$, that is used to measure and compare the performance of different models, is the accuracy of the model. The average value of the metrics over a number of experiments is denoted as $\overbar{\agreement}$ and $\overbar{\performance}$.

\section{Proposed method}\label{sec:method}

\subsection{Outline}

The idea is then to start from a DT whose training hyperparameters have been tuned (\eg using CV). The structure of the trained DT is transferred to an NDT with a parameter set $\theta$, as shown in \Fig{fig:outline}, including the batch size, the number of epochs, optimizer, the $\gamma_1$ and $\gamma_2$ values. During the procedure, the latter two are the only hyperparameters of the NDT that are not fixed. Therefore, finding $\theta^*$ in our case means finding the values of $\gamma_1$ and $\gamma_2$ for which the best average performance is observed.

$\gamma_1$ and $\gamma_2$ control the smoothness of the activation functions in the NDT: for higher values (\eg 100) the NDT behaves very similarly to the seed DT, while for lower values it gets closer to an NN with hyperbolic tangent activation functions. Varying progressively $\gamma_1$ and $\gamma_2$ from higher to lower values \emph{causes a progressive departure of the NDT model}: from the initial DT to an altered model that is relaxed and closer to an NN model. It is then possible to detect the point where better performance is obtained with respect to $\performance$, and to measure by using $\agreement$ how far from a DT is the best NDT model obtained with $\theta^*$, which is also written as $NDT^* := NDT(\theta^*)$. 
 
The core of the proposed method is then the search for $\theta^*$ and its subsequent interpretation, as well as that of $\performance(NDT^*)$ and $\agreement(NDT^*, DT)$.

\subsection{Relationship between $\gamma_1$ and $\gamma_2$}

In order to simplify the procedure and to provide easy to interpret results, we link the value of $\gamma_2$ to that of $\gamma_1$, such that $\gamma_2 = f(\gamma_1)$.
 
In \cite{Biau2018} it is suggested to use $\gamma_2 \ll \gamma_1$, since a smoother activation function in the second hidden layer allows for stronger weight corrections in the first hidden layer during backpropagation. 

On the other hand, a very low $\gamma_2$ value would make the function $\sigma_2:x \mapsto \tanh(\gamma_2 x)$ approach the flat zero-function. Since in our experiments $\gamma_1 \in [0.1, 900]$, we set the reasonable lower bound for $\gamma_2$ at 0.05.

We tested different functions on the datasets presented in \Tab{tab:datasets} and for different values of $\gamma_1$ using CV. The functions were: 
\begin{itemize}[topsep=0pt]
\item $x\mapsto x$ 
\item $x\mapsto\sqrt{x}$ 
\item $g:x\mapsto\log(10^{0.05}+x)$
\item $h:x\mapsto g(g(x))$.
\end{itemize}
Overall, $h$ was the function for which the NDTs achieved the best performance most of the times, across datasets and $\gamma_1$ values. 
 
In what follows, we use $\gamma$ to actually refer to $\gamma_1$ and then internally $\gamma_2$ is computed by:  
\begin{equation}
  \gamma_2 = h(\gamma_1) = \log\left(10^{0.05}+\log(10^{0.05}+\gamma_1)\right).
  \label{eq:gammas}
\end{equation}

\subsection{The algorithm of the procedure}

\Eq{eq:gammas} reduces the process of finding the $\theta^*$ to that of just estimating the $\gamma^*$. As we are interested in a progressive departure from any given DT, we can in fact produce multiple seed DTs, and for each of those to build several NDTs by decreasing gradually the value of $\gamma$. The $36$ tested values are in the ordered set $\Gamma =  \{900, 800, \mydots, 100\} \cup \{90, 80, \mydots, 10\} \cup \{9, 8, \mydots, 1\} \cup \{0.9, 0.8, \mydots, 0.1\}$.

For each trained NDT we measure the chosen performance metric $\performance$ and the agreement $\agreement$ (see \Sec{sec:metrics}). The overall procedure is detailed in \Alg{alg:method}. 

\begin{algorithm}[t]

\SetAlgoLined
  \KwIn{a dataset $D$, a set $\Gamma$ of values to test for $\gamma$}
  \KwOut{the value of $\gamma$ for which the highest mean performance is achieved, the agreement $\agreement$ with the reference DT and the performance $\performance$ of the corresponding NDT}
  find the hyperparameters $h$ of a DT model for the dataset

  \For{$i=1,...n$ \emph{times}}{
    form a training set, a test set, and a validation set from $D$

    build the new $DT_i := DT_i(h)$ with the training set

    measure $\performance(DT_i)$ on the test set

    \ForEach{$\gamma \in \Gamma$}{

      initialize the $NDT_{i,\gamma} := NDT(\gamma)$ using the $DT_i$ 
      
      fit the $NDT_{i,\gamma}$ using the training set and regularize it using the validation set

      measure $\agreement(NDT_{i,\gamma}, DT_i)$ and $\performance(NDT_{i,\gamma})$ on the test set
      
    }
  }
  
  $\overbar{M_{DT}} = \frac{1}{n}\sum_{i=1}^n \performance(DT_i)$
  
  $\overbar{M} = \left[\frac{1}{n}\sum_{i=1}^n \performance(NDT_{i, \gamma})\right]_{\gamma \in \Gamma}$
  
  $\overbar{A} = \left[\frac{1}{n}\sum_{i=1}^n \agreement(NDT_{i, \gamma}, DT_i)\right]_{\gamma \in \Gamma}$
  
  $\gamma^* =  \argmax_{\gamma \in \Gamma}\  \overbar{M}_\gamma$

  \Return{$\gamma^*$, $\overbar{A}$, $\overbar{M}$, $\overbar{M_{DT}}$}

  \caption{Pseudocode of the proposed method}
  \label{alg:method}
  \end{algorithm}

\section{Experiments}\label{sec:exps}

\subsection{Datasets}

The proposed method was tested on $5$ datasets, $1$ synthetic and $4$ containing real data taken from the UCI repository. Some of these datasets were chosen because either DTs or NNs are better suited to model them. 

First, we mention the synthetic dataset we use: 

the \dset{sim\_1000\_3} contains two classes, which are not linearly separable, and the data instances for each of them are generated by a normal distribution of different mean. For this dataset, we expect that an NDT with a low $\gamma^*$ will perform the best.

Following the same selection principle for real data, the Gastrointestinal Lesions in Regular Colonoscopy dataset (\dset{lesions}) was chosen as it is in high dimensions where DTs can be better in selecting only the most informative features \cite{Brown1993}. The \dset{mushroom} dataset was selected because it can be accurately modeled using simple rules. 
The rest of the real datasets offer points of comparison for when the more suitable model family is not as clear as above. They are the Spambase (\dset{spam}) and the Student Performance (\dset{student-math}) datasets. The characteristics of the datasets are given in \Tab{tab:datasets}.

\subsection{Experimental pipeline}

For each dataset, we first used CV to determine the depth of the DTs to be used (reported in \Tab{tab:datasets}), which was in fact the hyperparameter with the biggest impact on DT performance. Aside from removing the categorical variables (except for mushroom, where all its variables are categorical and were encoded) and the instances with missing data, no other preprocessing preceded.

For the NDTs, we set the number of epochs to $100$ for all datasets and the batch size was according to the size of each of them. We decided to use the Adam optimizer with default parameters and for regularization we use early stopping with a patience of $20$ epochs.

\begin{table*}[t]

  \begin{minipage}{1\textwidth}
      \centering
      \begin{tabular}{l||r|r|r|r|r|r|r|c|r}
        
        \toprule
         \textbf{Dataset}  &  \textbf{\#Features} &
        \textbf{\#Samples} & \textbf{DT depth} & \bm{$\gamma^*$}  & \bm{$\performance(DT)$}
        & \bm{$\performance(NDT(\gamma^*))$}
        & \bm{$\performance$} \textbf{diff.}
        & \textbf{Impr.} & \bm{$\agreement$}   \\
        \midrule
         \dset{sim\_1000\_3}       & 3    & 1000 & 4 & 9     & 0.906 (0.008)        & 0.989 (0.007)                   & -0.083 (0.010)       & \checkmark        & 0.819 (0.019)   \\
         \dset{mushroom}           & 23   & 8124 &  5   & 9     & 0.983 (0.002)        & 1.000 (0.000)                       & -0.016 (0.002)      & \checkmark        & 0.967 (0.003)   \\
         \dset{student-math}       & 31   & 395  &  4   & 200   & 0.733 (0.012)        & 0.725 (0.030)                    & 0.008 (0.030)        &        & 0.851 (0.140)    \\
         \dset{spam}               & 58   & 4601 &  5   & 0.6   & 0.916 (0.004)        & 0.940 (0.005)                    & -0.024 (0.007)      & \checkmark        & 0.842 (0.016)   \\
         \dset{lesions}            & 699  & 152  &  4   & 700   & 0.891 (0.025)        & 0.807 (0.080)                    & 0.084 (0.077)       &        & 0.448 (0.369)   \\
      \bottomrule
      \end{tabular}
      \caption{Summary of the experimental results. The results are the average of $30$ iterations and `Impr.' indicates whether the NDT achieved (on average) an improvement over the performance of the reference DT.}
      \label{tab:datasets}
  \end{minipage}
  \vspace{-1em}

  \end{table*}
To generate the training, validation, and test sets we use repeated random sub-sampling with proportions 50\%/25\%/25\% respectively, which we repeat $30$ times (so $n = 30$). We use stratification to ensure that classes are in the same proportions in the sets.

\begin{figure}[t]
\vspace{-1em}
  \centering
   
  \subfigure[\dset{sim\_1000\_3} dataset -- As the boundaries are getting relaxed, the NDT performs better, but agrees less with the reference DT. This suggests that a more flexible model might be more adapted for this dataset.]{\centering
  \includegraphics[width=0.9\columnwidth, viewport=10 10 450 335,clip]{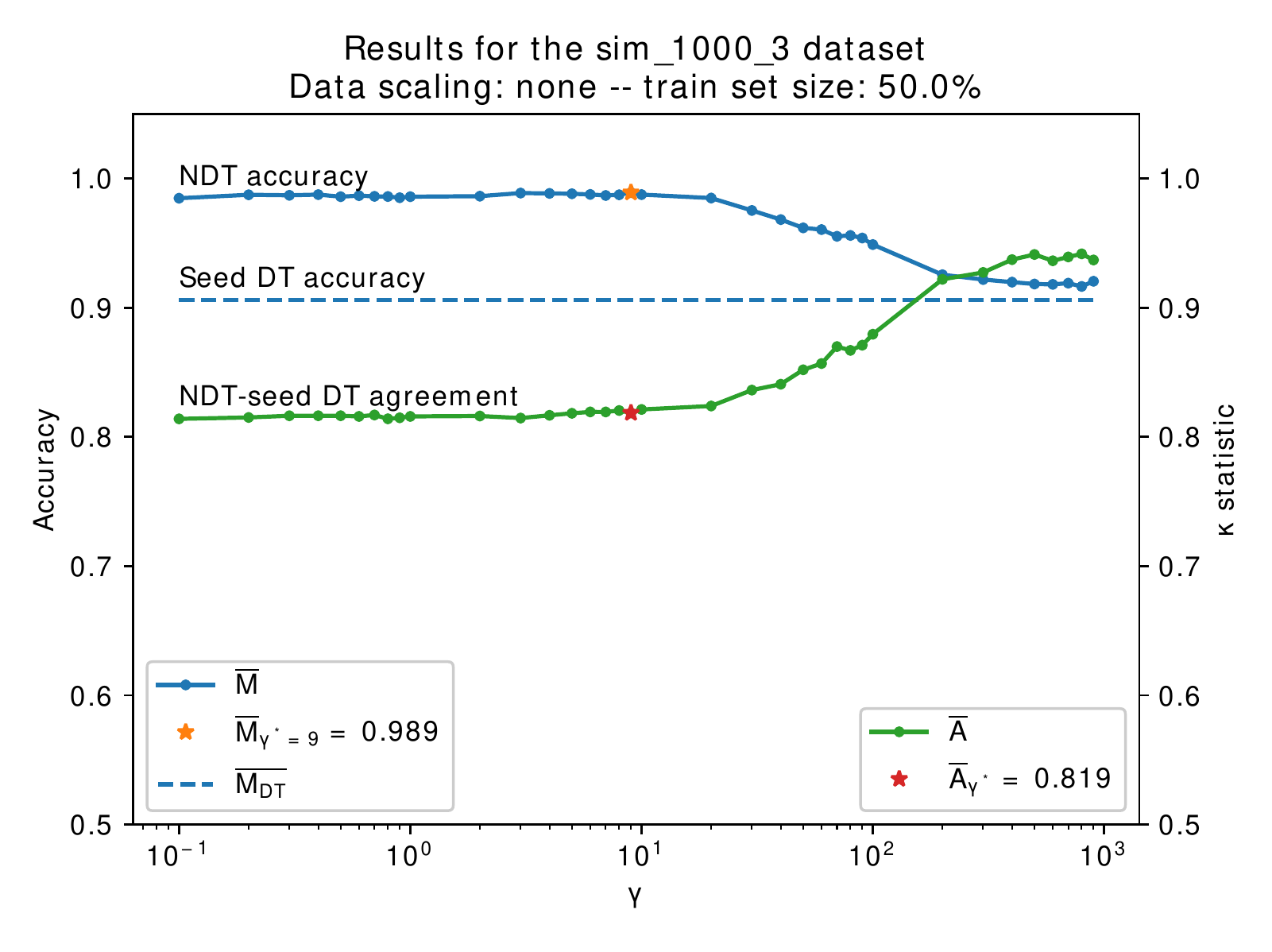}
  \vspace{-1.8em}
  \label{fig:results}
  }
  
    \subfigure[\dset{lesions} dataset -- Both the agreement and the accuracy of the NDT decrease with the value of $\gamma$, meaning a relaxation of the DTs boundaries is not beneficial in this case.]{\centering
    \includegraphics[width=0.9\columnwidth, viewport=10 10 450 335,clip]{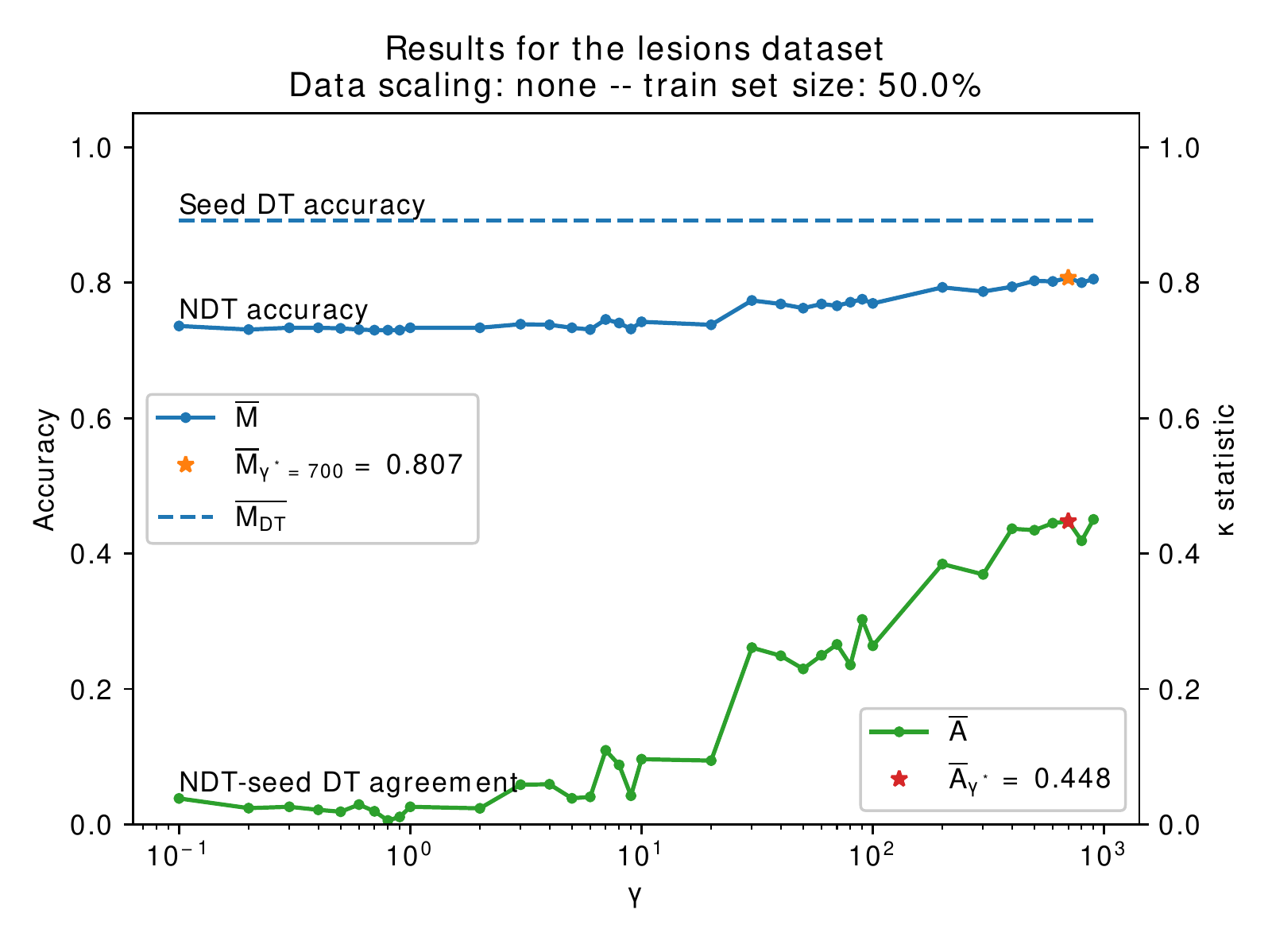}
    \label{fig:results_lesions}
    }

    \caption{Progressive relaxation (right-to-left across x-axis) of the NDT's decision boundaries with respect to its seed DT on two datasets.}
\end{figure}

\subsection{Results}

The method outputs the $\gamma^*$, $\performance(NDT^*)$, and $\agreement(NDT^*)$. These values are then to be interpreted to decide which model family is worth to further explore subsequently. We present here detailed examples of this interpretation for the \dset{sim\_1000\_3} and \dset{lesions} datasets.

\Fig{fig:results} shows the average
performance of the NDT with respect to the $\gamma$ value, and compares it to that of the initial DT. It tells us that, as expected, for higher values of $\gamma$ the NDT has an average performance that is close to that of its seed DT, and as $\gamma$ decreases, the accuracy of the NDT varies. In this case, the accuracy increases monotonically and reaches its maximum for $\gamma = 9$. For this value, $\sigma$
is closer to a $\tanh$ function and so it is a point where the NDT is close to a classic NN. However, the agreement between $NDT(\gamma^*)$ and the DT is high (0.819, see \Tab{tab:datasets}). 
This means that a DT and an NN model may not differ much in their decisions, nonetheless if one is interested in obtaining better performance at the cost of sacrificing the explainability, then further exploring of the the NN family is the most promising direction to go.

\Fig{fig:results_lesions} shows both the performance and the agreement decreasing as the value of $\gamma$ decreases. Here the best average accuracy for the NDT is reached when $\gamma^*=700$. In this case the function $\sigma$ is very close to a threshold function, and so the NDT behaves similarly to a DT. In contrast, there is no improvement over the seed DT, even for $NDT^*$, and even for high values of $\gamma$ the $\overbar{A}$ is fairly low. This is due to the setting of $\gamma_2$ according to \Eq{eq:gammas}. Indeed, an NDT behaves most like its seed DT when \emph{both} $\gamma_1$ and $\gamma_2$ are high. These factors taken into account, our interpretation of these results would be that exploring the DT model family further might be more propitious.

Similar reasoning can be developed for the rest of the datasets using the results reported in \Tab{tab:datasets}. Similar graphs can be drawn for all the datasets, however, their interpretation might be less straightforward than for the example of \Fig{fig:results}. It is worth noting though, that the preprocessing applied to the datasets can greatly affect the results.

\section{Conclusion}\label{sec:conclusion}

The method introduced in this work allows us to determine if more flexible models can outperform a trained DT on a given learning task. This is done in a controlled way by progressively relaxing DTs' decision boundaries. Furthermore, the agreement metric provides insight about how far it was necessary to depart from the reference model in order to achieve an improvement in performance (if there is one). This can be useful when deciding between a rigid but more explainable model and a more flexible but less interpretable one. Our method's starting point is a trained DT.
Nonetheless, the optimization of $\gamma_1$ and $\gamma_2$ remains an open problem.
Extending this work could be attempted in the direction of enriching the experimental results and comparisons to other methods.

\bibliographystyle{acm}
\bibliography{mfs_short}

\end{document}